\documentclass{article}
\usepackage{spconf,amsmath,graphicx}

\usepackage{amsfonts}
\usepackage{caption}
\usepackage[labelformat=simple]{subcaption}
\usepackage{float}
\usepackage{amssymb}
\usepackage{enumitem}
\usepackage{algorithm} 
\usepackage{algorithmic}
\usepackage[algo2e]{algorithm2e}
\usepackage{amsthm,mathtools}
\usepackage{url}
\usepackage{stmaryrd}
\usepackage{cite}
\usepackage{multirow}
\usepackage[table,xcdraw]{xcolor}
\usepackage{url}
\usepackage{ragged2e}
\usepackage{epstopdf}

\usepackage{booktabs}
\usepackage{makecell}
\usepackage{threeparttable}

\DeclareMathOperator{\diag}{\operatorname{diag}}

\DeclareMathOperator{\softmax}{\operatorname{softmax}}
\DeclareMathOperator{\relu}{\operatorname{ReLU}}

\title{Inductive Graph Neural Networks for Moving Object Segmentation}
\name{Wieke Prummel$^{1}$, Jhony H. Giraldo$^{2}$, Anastasia Zakharova$^{1}$, Thierry Bouwmans$^{1}$}
\address{$^{1
}$ Laboratoire Mathématiques, Image et Applications (MIA), La Rochelle Université, France \\
$^{2}$ LTCI, Télécom Paris - Institut Polytechnique de Paris, France}
\begin{document}
%
\maketitle
\begin{abstract}

{Moving Object Segmentation (MOS) is a challenging problem in computer vision, particularly in scenarios with dynamic backgrounds, abrupt lighting changes, shadows, camouflage, and moving cameras. While graph-based methods have shown promising results in MOS, they have mainly relied on transductive learning which assumes access to the entire training and testing data for evaluation. However, this assumption is not realistic in real-world applications where the system needs to handle new data during deployment. In this paper, we propose a novel Graph Inductive Moving Object Segmentation (GraphIMOS) algorithm based on a Graph Neural Network (GNN) architecture. Our approach builds a generic model capable of performing prediction on newly added data frames using the already trained model. GraphIMOS outperforms previous inductive learning methods and is more generic than previous transductive techniques. Our proposed algorithm enables the deployment of graph-based MOS models in real-world applications.}
\end{abstract}
\section{Introduction}
\label{sec:intro}


{Moving Object Segmentation (MOS) is an important problem in computer vision, particularly in surveillance system applications \cite{garcia2020background}. The goal of MOS is to identify and separate the pixels or regions in a video that correspond to moving objects from the static background or other static objects. Deep learning models \cite{BOUWMANS20198} have demonstrated strong performance on large-scale datasets. However, as the quality of data improves, these models become increasingly complex and computationally intensive, even with fast algorithms \cite{hou2020super}, few-shot learning methods \cite{9897894}, and specialized architectures \cite{9191151}.}

\begin{figure}
    \centering
    \includegraphics[width=\columnwidth]{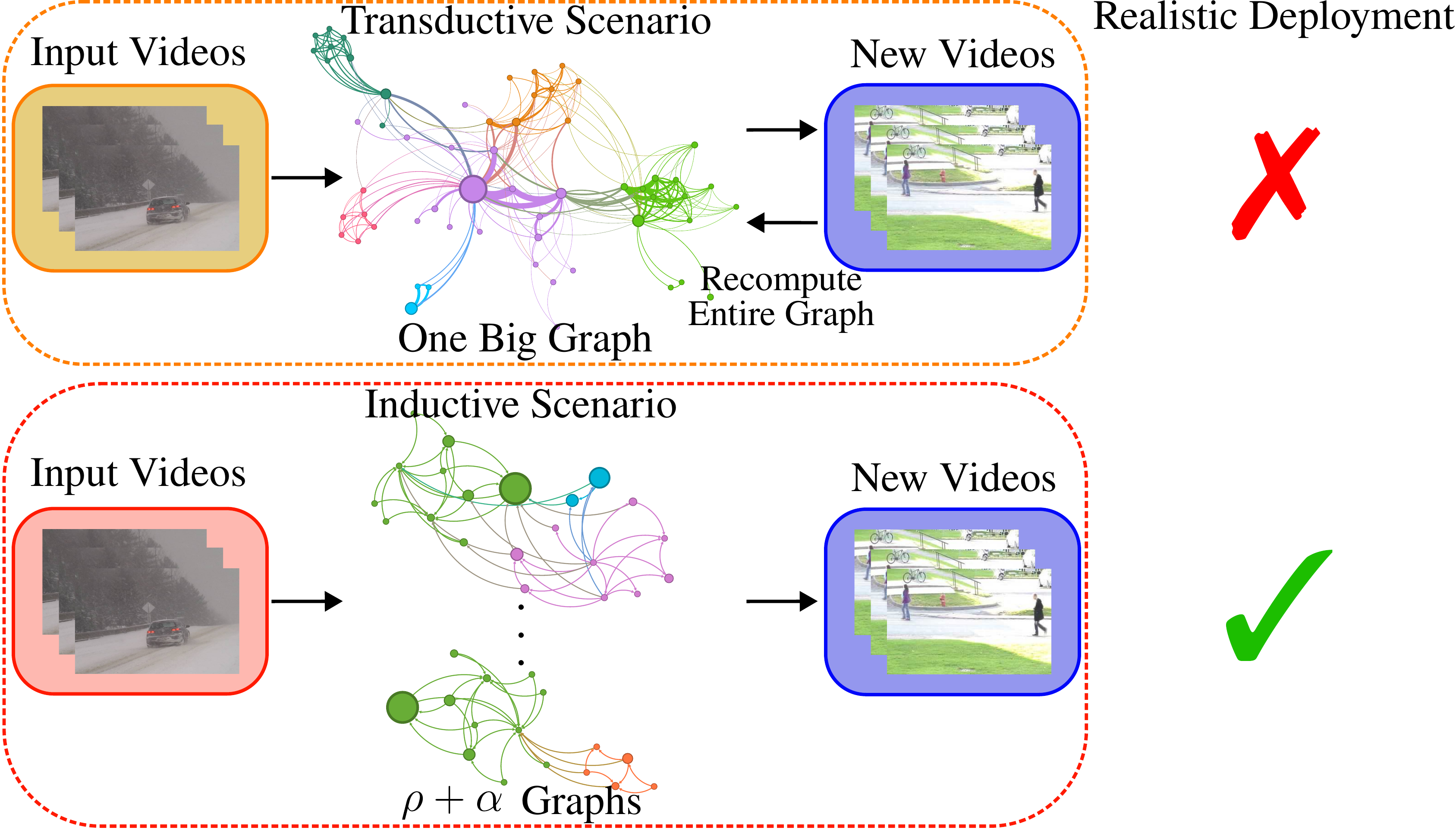}
    \caption{Transductive vs. inductive learning on MOS}
    \label{fig:teaser}
\end{figure}

{The most common deep models are mostly supervised and can be divided into four groups \cite{hou2023survey}: 2D Convolutional Neural Networks (CNNs) \cite{braham2016deep}, 3D CNNs \cite{akilan20193d}, transformer neural networks \cite{9607687}, and generative adversarial networks \cite{bakkay2018bscgan,SULTANA2022240}. In addition to these, some state-of-the-art (SOTA) techniques have been combined with deep methods to create novel approaches, such as MotionRec \cite{mandal2020motionrec}, RT-SBS \cite{cioppa2020real}, and GraphMOS \cite{giraldo2020graph}. Recent graph-based algorithms like GraphMOS \cite{giraldo2020graph} and GraphMOD-Net \cite{10.1007/978-3-030-81638-4_3} use semi-supervised learning posed as a graph signal reconstruction problem. These methods are inspired by the theory of graph signal processing \cite{ortega2018graph} and have shown promising results.
However, these graph-based methods are transductive in nature, meaning that the model needs to be fully retrained and the graph regenerated whenever a new video is added. To address this issue, we propose inductive techniques for graph-based MOS, where multiple graphs are built instead of a single large graph. This approach reduces the need for rebuilding the whole graph and retraining the model, making it more suitable for real-world deployments, as shown in Fig. \ref{fig:teaser}.}

\begin{figure*}
    \centering
    \includegraphics[width=0.98\textwidth]{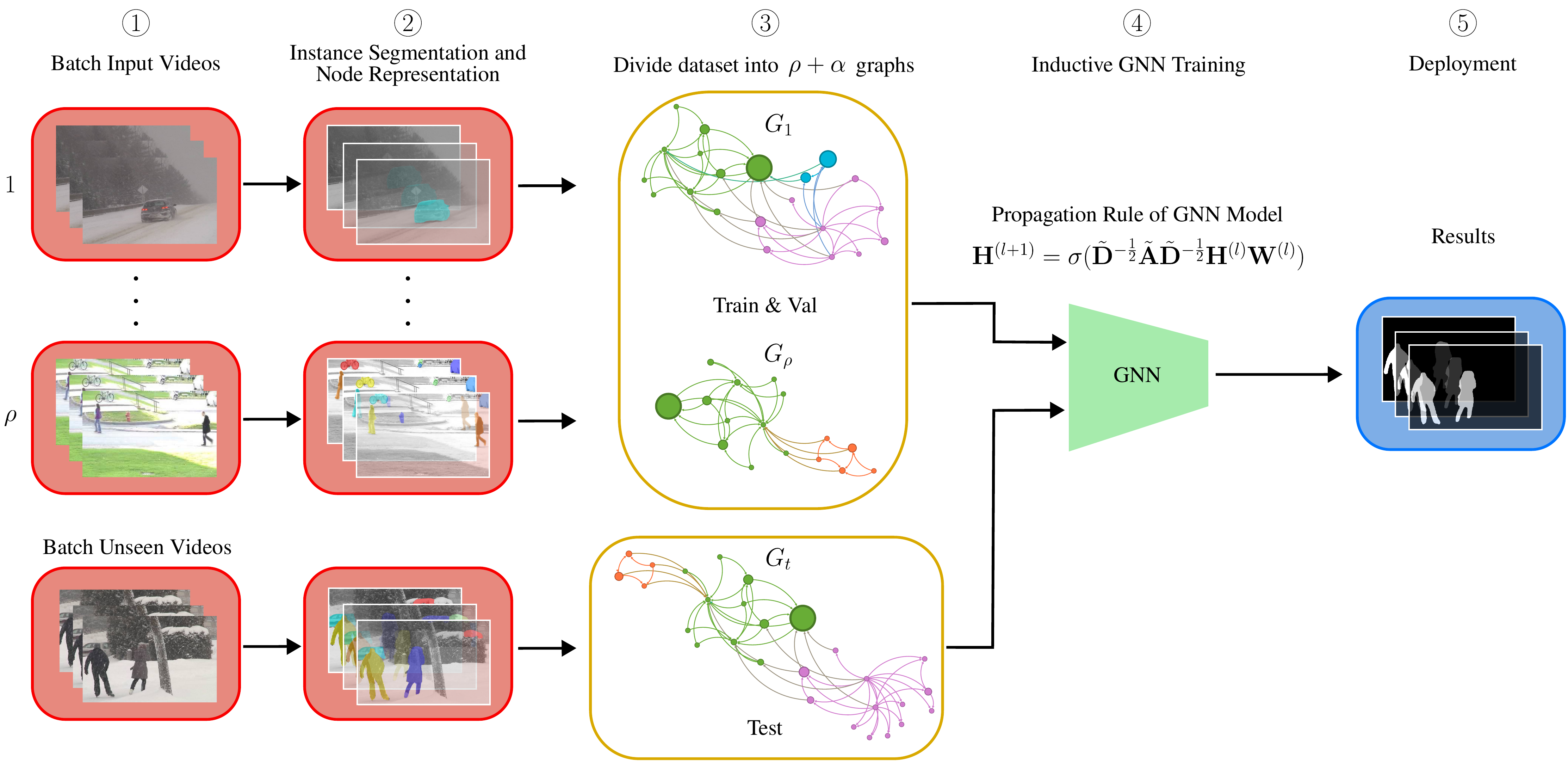}
    \caption{Pipeline of the graph inductive moving object segmentation algorithm via graph neural network learning. After instance segmentation and node feature extraction, the dataset is divided into $\rho$ graphs for training-validation and into $\alpha$ testing graphs. The algorithm then classifies nodes in the graphs as either moving or static objects.}
    \label{fig:pipeline}
\end{figure*}


In this work, we propose a novel Graph Inductive Moving Object Segmentation (GraphIMOS) algorithm based on Graph Neural Networks (GNNs) \cite{bronstein2021geometric}. We represent each instance of the video as a node in a graph, which is generated using a Mask Region Convolutional Neural Network (Mask R-CNN) with a ResNet-50 and Feature Pyramid Network (FPN) backbone. We use optical flow, intensity, and texture features to represent the nodes. Instead of creating a single large graph, as in \cite{giraldo2020graph}, we generate $\rho + \alpha$ $k$-Nearest Neighbors ($k$-NN) graphs, where each graph can have a different number of nodes. We set $\rho = 3$ in the experiments for the training and validation graphs, which are then fed to the proposed GNN model for training and hyperparameter optimization. To evaluate the performance of our model, we test it on one single graph $\alpha = 1$, built from previously unseen videos. Our approach is evaluated on the challenging Change Detection 2014 (CDNet 2014) dataset \cite{wang2014cdnet}, and we demonstrate competitive results against previous inductive methods.


This paper makes two main contributions. Firstly, we propose an inductive graph-based framework for Moving Object Segmentation (MOS), which is an important step toward the real-world deployment of graph-based methods in video surveillance applications. Secondly, we introduce a novel GNN architecture specifically designed for MOS, called GraphIMOS. To the best of our knowledge, GraphIMOS is the first graph-based inductive learning approach for MOS. The remainder of this paper is organized as follows. In Section 2, we introduce the preliminary concepts and describe the GraphIMOS algorithm and architecture. Section 3 presents the experimental setup and results. Finally, in Section 4, we present our conclusions.


\section{Graph Inductive Moving Object Segmentation}

This section presents the preliminaries of this paper and the proposed inductive graph-based MOS method.
Figure \ref{fig:pipeline} shows the pipeline of GraphIMOS, which consists of instance segmentation, node representation, graph construction, and inductive GNN training and evaluation.


\subsection{Preliminaries}

A graph is a mathematical entity that can be represented as $G=(\mathcal{V}, \mathcal{E})$, where $\mathcal{V}=\{1,2,\dots,N\}$ is the set of nodes and ${\mathcal{E}\subseteq \{(i,j)\mid i,j\in \mathcal{V}\;{\textrm {and}}\;i\neq j\}}$ is the set of edges connecting the nodes $i$ and $j$.
The adjacency matrix $\mathbf{A} \in \mathbb{R}^{N \times N}$ is a popular choice of shift operator in GNNs \cite{kipf2016semi}, where $\mathbf{A}_{(i,j)} = a_{i,j}~\forall~(i,j) \in \mathcal{E}$, and $0$ otherwise.
For unweighted graphs we have that $\mathbf{A} \in \{0,1\}^{N \times N}$.
Similarly, we have the diagonal degree matrix given by $\mathbf{D}=\diag(\mathbf{A1})$.
In GNNs, we commonly associate a vector of features $\mathbf{x}_i \in \mathbb{R}^F$ to each node $i$.
Therefore, we can represent the whole set of input features as $\mathbf{X} = [\mathbf{x}_1, \mathbf{x}_2, \dots, \mathbf{x}_N]^{\mathsf{T}} \in \mathbb{R}^{N \times F}$.
In this paper, we use undirected and weighted graphs.

GraphIMOS is designed to be agnostic to the choice of the GNN used. However, in our experiments, we use the widely used Graph Convolutional Network (GCN) \cite{kipf2016semi}. The propagation rule of GCN is given by:
\begin{equation}
    \mathbf{H}^{(l+1)}=\sigma(\tilde{\mathbf{D}}^{-\frac{1}{2}}\tilde{\mathbf{A}}\tilde{\mathbf{D}}^{-\frac{1}{2}}\mathbf{H}^{(l)}\mathbf{W}^{(l)}),
    \label{eqn:propagation_rule_GCN}
\end{equation}
where $\tilde{\mathbf{A}}=\mathbf{A}+\mathbf{I}$, $\tilde{\mathbf{D}}$ is the degree matrix of $\tilde{\mathbf{A}}$, $\mathbf{H}^{(l)}$ is the output matrix of layer $l$ (with $\mathbf{H}^{(0)}=\mathbf{X}$), $\mathbf{W}^{(l)}$ is the matrix of trainable weights in layer $l$, and $\sigma(\cdot)$ is an activation function such as $\relu$ or $\softmax$. It is worth noting that \eqref{eqn:propagation_rule_GCN} reduces to the regular multi-layer perceptron when $\tilde{\mathbf{D}}^{-\frac{1}{2}}\tilde{\mathbf{A}}\tilde{\mathbf{D}}^{-\frac{1}{2}} = \mathbf{I}$.

\subsection{Segmentation and Node Representation}

For the region proposal, we use Mask R-CNN with ResNet-50 as the backbone and construct graphs using a $k$-NN method. The feature representation proposed in \cite{giraldo2020graph} is utilized, where the outputs generated by Mask R-CNN are represented as nodes in the graphs. These instances are associated with meaningful representations, such as optical flow, intensity, and texture features. Finally, all the features are concatenated to form a $930$-dimensional vector, which represents each instance. For further details on the feature extraction process, please refer to \cite{giraldo2021graphbgs, giraldo2020graph}.

\subsection{Graph Mini-batch}



The proposed framework is distinct from previous graph-based MOS methods because it is an inductive architecture. For instance, in \cite{giraldo2020graph}, the whole graph would need to be rebuilt every time a new video is fed into the algorithm.
This requires the optimization problem to be solved again, making deployment in real-world scenarios more challenging. 
Therefore, instead of creating a single graph from the data, our proposed approach creates $\rho + \alpha$ separate graphs, each with a different number of nodes. This ensures that the data in the $\rho$ training-validation graphs are not connected. The adjacency matrices are arranged in a block-diagonal manner to construct a comprehensive graph that includes multiple distinct subgraphs as follows:
\begin{equation}
  \mathbf{A} =
  \begin{bmatrix}
    \mathbf{A}_{1} & & \\
    & \ddots & \\
    & & \mathbf{A}_{\rho}
  \end{bmatrix},
\end{equation}
where $\mathbf{A}$ is the adjacency matrix in the mini-batch.
We define $\rho$ for the number of training and validation graphs, and $\alpha$ for the testing graphs as shown in Fig. \ref{fig:pipeline}. Our adjacency matrices are stored efficiently, using a sparse representation that only keeps track of non-zero entries: the edges. This means that there is no extra memory overhead. The node and target features are integrated into the node dimension through simple concatenation.


\subsection{GNN Architecture}


Our model consists of two GCNConv layers \cite{kipf2016semi}, various ReLU activation layers, and three linear layers. To reduce overfitting and improve generalization on unseen videos, we use five dropout and four pair batch normalizations (PairNorm) \cite{zhao2020pairnorm}. These techniques also enhance the training stability of the model and prevent over-smoothing. PairNorm is particularly effective in achieving a better convergence of the proposed model. Our goal is to learn meaningful representations of the graph structure and node features to classify objects as either static or moving. The final layer of our model employs a log $\softmax$ function defined as follows: $\log\_\softmax = \log(\frac{1}{s}\exp{x_{i}})$ where $s = \sum_{i}\exp{x_{i}}$. This GNN architecture enables GraphIMOS to be easily deployed in real-world applications.



\begin{table}
\caption{Data Partitioning.}
\centering
\resizebox{\columnwidth}{!}{
\begin{tabular}
{lcccc} 
\toprule
\textbf{Sequences} & \textbf{Training} & \textbf{Validation} & \textbf{Testing}\\
\midrule
S2, S3, S1, S4 (Exp1) & $G2,G3$ & $G1$ & $G4$ \\
S1, S3, S4, S2 (Exp2) & $G1,G3$ & $G4$ & $G2$ \\
S2, S3, S4, S1 (Exp3) & $G2,G3$ & $G4$ & $G1$ \\
S1, S2, S4, S3 (Exp4) & $G1,G2$ & $G4$ & $G3$ \\
\bottomrule 
\label{table:Data partition}
\end{tabular}
}
\end{table}

\begin{table}
\caption{F-Measure Experiments.}
\centering
\resizebox{\columnwidth}{!}{
\begin{tabular}
{lcccc} 
\toprule 
\textbf{F-Measure} & \textbf{Exp1} & \textbf{Exp2} & \textbf{Exp3}  & \textbf{Exp4}\\
\midrule 
F-Measure validation & $0.8420$ & $0.8006$ & $0.8386$ & $0.8014$ \\
F-Measure test & $0.8388$ & $0.7969$ & $0.8367$ & $0.7567$\\

\bottomrule 
\label{table: F-measure}
\end{tabular}
}
\end{table}

\section{Experiments and Results}

This section introduces the metrics, and the dataset graph partitioning used to conduct the different experiments on GraphIMOS. 


\subsection{Evaluation Metrics}


The evaluation metrics F-Measure, precision, and recall are defined as follows:
\begin{gather}
    \nonumber
    \text{Recall} = \frac{\text{TP}}{\text{TP}+\text{FN}}, \text{ } \text{Precision} = \frac{\text{TP}}{\text{TP}+\text{FP}},\\
    \label{eqn:f-measure}
    \text{F-measure} = 2\frac{\text{Precision}\times \text{Recall}}{\text{Precision}+\text{Recall}},
\end{gather}
where TP, FN, and FP are the true positives, false negatives, and false positives, respectively. According to Table \ref{table:Data partition}, we calculate the F-Measure in Table \ref{table: F-measure} on node level classification.

\begin{table*}[h]
\caption{Average F-Measure for transductive and inductive methods. The best score of all methods appears in bold, and the best score of the inductive methods is bold underlined. The columns contain the CDNet2014 \cite{wang2014cdnet} challenges : bad weather (BWT), baseline (BSL), camera jitter (CJI), dynamic background (DBA), intermittent object motion (IOM), low frame rate (LFR), PTZ, shadow (SHW), and thermal (THL).} 
\centering
\resizebox{0.95\textwidth}{!}
{
\begin{tabular}{lcccccccccc}
\toprule 
\textbf{Method} & \textbf{BSL} & \textbf{BWT} & \textbf{IOM} & \textbf{LFR} & \textbf{PTZ} & \textbf{THL} & \textbf{CJI} & \textbf{SHW} & \textbf{DBA} & \textbf{Overall} \\
\midrule 
\multicolumn{11}{c}{\textbf{Transductive Learning Methods}} \\
GraphMOS \cite{giraldo2020graph} & ${0.9398}$ & ${0.8294}$ & $0.3607$ & $\textbf{0.5538}$ & ${0.7599}$& $\textbf{0.7292}$ & ${0.7005}$ &  $\textbf{0.9653}$ & ${0.7334}$ & ${0.7302}$ \\
GraphMOD-Net \cite{giraldo2021graph} (Original) & $\textbf{0.9550}$ & $\textbf{0.8390}$ & $\textbf{0.5540}$ & $0.5210$ & $\textbf{0.7700}$ & $0.6820$ & $\textbf{0.7200}$ & $0.9420$ & $\textbf{0.8510}$ & $\textbf{0.7593}$ \\
\midrule 
\multicolumn{11}{c}{\textbf{Inductive Learning Methods}} \\
FgSegNet \cite{lim2020learning} & $0.5641$ & $0.2789$ & $0.3325$ & $0.2115$ & $0.1400$ & $0.3584$ & $ 0.2815$ & $0.3809$ & $0.2067$ & $0.3061$ \\
GraphMOD-Net (Modified) & $0.6474$ & $0.6268$ & $0.5243$ & $0.5337$ & $0.5899$ & $0.5484$ & $0.4926$ & $0.6587$ & $\textbf{\underline{0.6254}}$ & $0.5831$ \\
GraphIMOS (Ours) & $\textbf{\underline{0.7003}}$ & $\textbf{\underline{0.6377}}$ & $\textbf{\underline{0.5284}}$ &  $\textbf{\underline{0.5478}}$ & $\textbf{\underline{0.5932}}$  &  $\textbf{\underline{0.6453}}$ &  $\textbf{\underline{0.6700}}$ &  $\textbf{\underline{0.6807}}$ & $0.5868$ & $\textbf{\underline{0.6211}}$ \\
\bottomrule 
\end{tabular}
}
\label{table: Average F-measure}
\end{table*}

\begin{figure*}
    \def\myim#1{\includegraphics[width=0.12\textwidth]{#1}}
    \centering
    \setlength\tabcolsep{1 pt}
    \renewcommand{\arraystretch}{0.2}
    \begin{tabular}{lccccc}
    \toprule 
    \textbf{CDNet 2014} & \textbf{Original} &\textbf{Ground Truth} &\textbf{FgSegNet} & \textbf{GraphMOD}& \textbf{GraphIMOS} \\
    \midrule
    Baseline, pedestrians & \makecell[l]{\myim{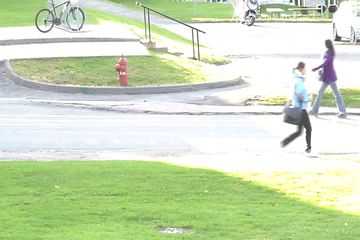}} & \makecell[l]{\myim{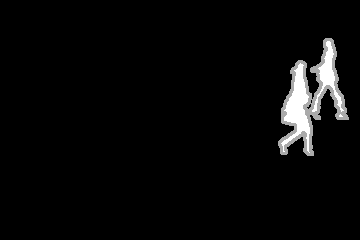}} & \makecell[l]{\myim{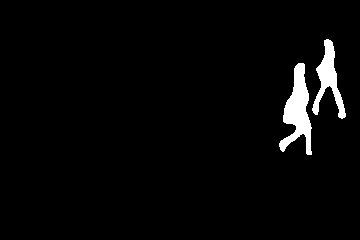}} & \makecell[l]{\myim{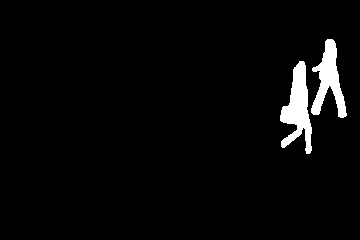}} & \makecell[l]{\myim{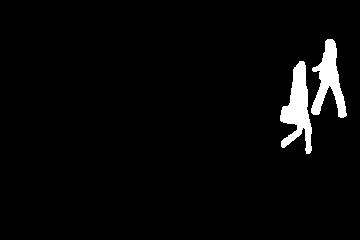}} \\
    Intermittent object motion, tramstop & \makecell[l]{\myim{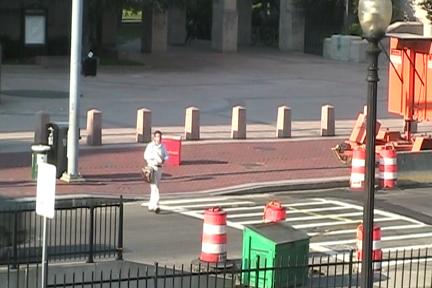}} & \makecell[l]{\myim{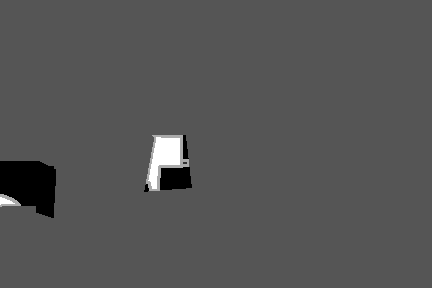}} & \makecell[l]{\myim{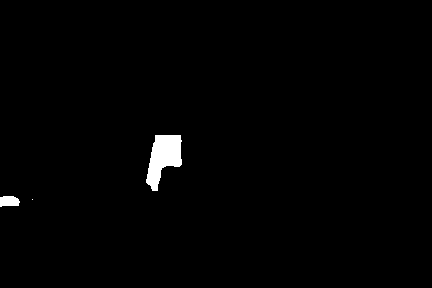}} & \makecell[l]{\myim{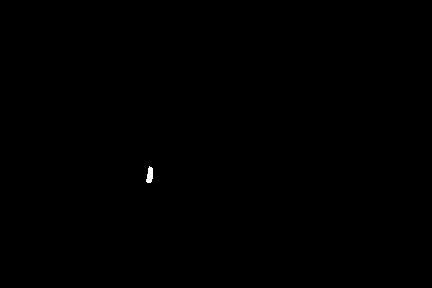}} & \makecell[l]{\myim{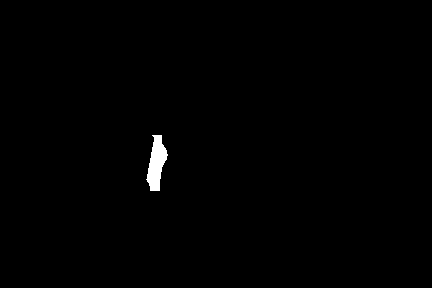}} \\
    Shadow, cubicle & \makecell[l]{\myim{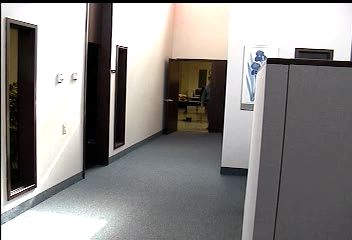}} & \makecell[l]{\myim{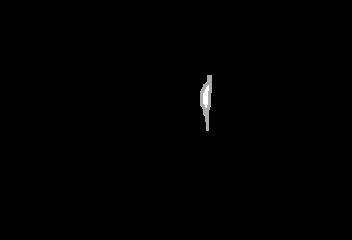}} & \makecell[l]{\myim{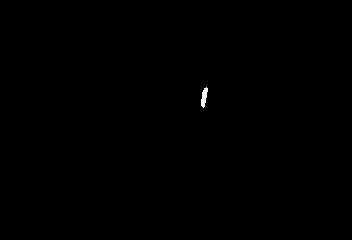}} & \makecell[l]{\myim{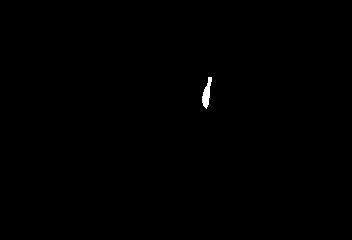}} & \makecell[l]{\myim{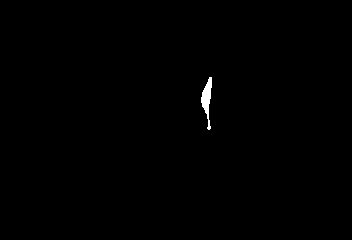}} \\
    \bottomrule
\end{tabular}
\caption{Comparisons of the visual results on unseen videos of the proposed GraphIMOS algorithm with two inductive methods: FgSegNet \cite{lim2020learning} and the modified GraphMOD-Net \cite{giraldo2021graph}. We show results from three different challenges from CDNet 2014 \cite{wang2014cdnet}.}
\label{fig:visual_results}
\end{figure*}

\subsection{Experiments}


We evaluate GraphIMOS against several SOTA algorithms using the large-scale CDNet 2014 dataset \cite{wang2014cdnet}. To build an inductive framework, we construct multiple graphs by dividing the data into four sequences: S1, S2, S3, and S4 as shown in Table \ref{table:Data partition}. Therefore, we run 4 experiments to compute the F-Measure in the whole dataset.
These sequences were chosen to enable unseen videos evaluation, following the approach in \cite{tezcan2021bsuv}. For each sequence, we compute the results and construct one graph per sequence : $G1$, $G2$, $G3$, and $G4$.


In each experiment, we use two graphs to train and evaluate GraphIMOS, and one graph for testing. For each experiment we chose to use a different graph to test the performance of the proposed algorithm on unseen videos.

\subsection{Implementation Details}

We employ the PyTorch Geometric library \cite{fey2019fast} to implement our proposed architecture. The $k$-nearest neighbors ($k$-NN) graphs are constructed with a value of $k=40$. In order to prevent overfitting, we use a dropout with a coefficient of $0.5$. For optimization, we use stochastic gradient descent (SGD) with a momentum of $0.9$, a learning rate of $0.01$, and $5e{-4}$ for weight decay. We train our model for a maximum of $500$ epochs, using a graph-batch size of $1$. Our model architecture includes $5$ hidden layers, and we opt for the negative log likelihood loss for training and evaluating the GNN model.



\subsection{Results and Discussion}

We compare GraphIMOS with FgSegNet \cite{lim2020learning}, GraphMOD-Net \cite{giraldo2021graph}, and GraphMOS \cite{giraldo2020graph}. While GraphMOD-Net is transductive in nature, we adapt it to our data partitioning and experimental framework for a fair comparison with GraphIMOS.
We also include GraphMOS and GraphMOD-Net with their original performances as references, since they are transductive techniques. The numerical and visual results of the compared methods are shown in Fig. \ref{fig:visual_results} and Table \ref{table: Average F-measure}, which demonstrate that GraphIMOS surpasses previous inductive learning methods. Our experiments also reveal that GraphMOD-Net \cite{giraldo2021graph} shows a performance degradation when evaluated in an inductive setting, possibly due to the challenges of real-world deployments. GraphIMOS strikes a better balance between performance and realistic deployment, making it a promising candidate for real-world applications.

\section{Conclusion}

This paper introduces GraphIMOS, a novel approach that uses GNNs, graph mini-batches, and inductive learning for MOS. The proposed algorithm consists of four key components: instance segmentation using Mask R-CNN, feature extraction for node representation, $k$-NN for graph construction, and a GNN-based inductive learning algorithm. To the best of our knowledge, GraphIMOS is the first approach that uses graph-based inductive learning for MOS, demonstrating its novelty and potential. Compared to previous works such as GraphMOD-Net, GraphIMOS offers improved performance and a better trade-off between performance and practical deployment. For future work, we plan to add skip connections and deeper GNN models.




\bibliographystyle{IEEEbib}
\bibliography{refs}

\end{document}